\documentclass[review]{elsarticle}

\makeatletter
\def\ps@pprintTitle{%
\let\@oddhead\@empty
\let\@evenhead\@empty
\def\@oddfoot{}%
\let\@evenfoot\@oddfoot}
\makeatother

\usepackage{amssymb}
\usepackage{hyperref}
\usepackage{xcolor}
\usepackage{graphicx}
\usepackage{url}
\usepackage{booktabs}
\usepackage{amsmath}   
\usepackage{multirow}  
\usepackage{array}  
\usepackage{orcidlink}
\usepackage{geometry}
\usepackage{lineno}

\newcommand{\Tri}{\mathrm{Tri}}
\newcommand{\triMF}[3]{\Tri\bigl(#1,#2,#3\bigr)}

\begin{document}

\begin{frontmatter}

\title{ff4ERA: A new Fuzzy Framework for Ethical Risk Assessment in AI}

\author[A]{Abeer Dyoub \corref{cor1}  \orcidlink{0000-0003-0329-2419}}\ead{abeer.dyoub@uniba.it} \cortext[cor1]{Corresponding Author}
\author[C]{Ivan Letteri \orcidlink{0000-0002-3843-386X}} \ead{ivan.letteri@univaq.it}
\author[A,B]{Francesca A. Lisi  \orcidlink{0000-0001-5414-5844}} \ead{FrancescaAlessandra.Lisi@uniba.it}

\affiliation[A]{organization={Department of Computer Science},
            addressline={University of Bari ``Aldo Moro''}, 
            city={Bari},
            country={Italy}}
\affiliation[B]{organization={Centro Interdipartimentale di Logica e Applicazioni (CILA)},
            addressline={University of Bari ``Aldo Moro''}, 
            city={Bari},
            country={Italy}}
\affiliation[C]{organization={Department of Life, Health and Environmental Sciences},
            addressline={University of L'Aquila}, 
            city={Coppito - L'Aquila},
            country={Italy}}

\begin{abstract}

\textbf{Background and Motivation:}
The emergence of Symbiotic AI (SAI) introduces new challenges to ethical decision-making, as it deepens human–AI collaboration. With increased symbiosis, AI systems pose greater ethical risks, including harm to human rights and trust. Ethical Risk Assessment (ERA) becomes a crucial step in guiding decisions that minimize such risks. However, ERA is hindered by inherent uncertainty, vagueness, and incomplete information. Furthermore, morality is context-dependent and imprecise.
This motivates the need for a flexible, transparent, yet robust framework for ERA.

\textbf{Objectives:}
This work aims to support ethical decision making by quantitatively assessing and prioritizing multiple ethical risks, so that artificial agents can choose actions aligned with human values and acceptable risk levels.

\textbf{Methodology:}
We introduce ff4ERA, a fuzzy framework that integrates Fuzzy Logic, Fuzzy Analytical Hierarchy Process (FAHP), and Certainty Factors (CF) to quantify possible ethical risks by calculating an ethical risk score (ERS) for each ethical risk type. The final ERS for each ethical risk is obtained by combining its FAHP‐derived weight, the propagated CF, and the risk level. The framework provides a robust mathematical approach for collaborative modeling of ERA and allows for a step by step analysis of ERA in a systematic manner.

\textbf{Results:}
The case study confirms that the proposed framework produces ethically meaningful and context-sensitive risk scores, reflecting both expert input and sensor-based evidence. Risk scores vary consistently with changes in relevant factors, while remaining robust to unrelated inputs. Local sensitivity analysis reveals predictable, mostly monotonic behavior across input perturbations. The global Sobol analysis highlights the dominant influence of expert-defined weights and certainty factors, validating the model’s structured design. Overall, the results demonstrate the framework’s ability to produce interpretable, traceable, and risk-aware ethical assessments.

\textbf{Conclusions:}
ff4ERA delivers explainable, robust ethical risk scores, enabling “what‑if” analyses and guiding designers to calibrate membership functions and expert judgments for reliable ethical decision support.

\end{abstract}


\end{frontmatter}



\section{Introduction}
\label{intro}

\paragraph{Background}

Symbiotic Artificial Intelligence (SAI) reimagines the role of AI as a cooperative partner that enhances human capabilities instead of replacing them. In SAI, intelligent systems learn from and adapt to users in real time, augmenting decision‑making and skill sets to achieve outcomes that neither humans nor machines could reach alone. As these human–AI partnerships deepen,  the consequences of AI actions become more significant, heightening both potential benefits and ethical risks. Consequently, SAI demands robust machine‑ethics solutions to safeguard human rights, build trust, and ensure that long‑term collaboration between humans and AI remains both safe and mutually rewarding.

In response, the European Union’s AI Act adopts a \emph{risk‑based} regulatory approach, classifying AI applications by their potential to cause harm and imposing corresponding obligations on providers and users \cite{EC2021}.  However, translating high‐level regulatory risk categories into concrete system design and runtime decision processes remains a major challenge.  
Existing machine‑ethics paradigms, ranging from rule‑based “top‑down” deontic logics \cite{Allen2005} to learning‑based POMDP and reinforcement‑learning frameworks \cite{Abel2016}, each address aspects of this problem but lack a unified, transparent methodology for quantifying and prioritizing multiple interacting ethical risks under uncertainty.

Anytime the actions/decisions of an AI-based system have potential to impact humans positively or negatively, it is a matter of ethical concern. In the ethical context, it is crucial to prevent AI-based systems from causing harm. The potential risk of causing harm of any kind to humans is what we refer to as 'ethical risk' in this paper. There are different categories of ethical risks involving different types of harm, some examples are:
\begin{itemize}
    \item physical harm (e.g. injury or death)
    \item mental harm (e.g. depression, anxiety, addiction)
    \item violation of autonomy
    \item violation of privacy or confidentiality
    \item violation of trust and respect
    \item violation of fairness (discrimination)
\end{itemize}

\paragraph{Motivation}
Autonomous and Symbiotic AI systems are being deployed in domains as sensitive as elder care, healthcare triage, and critical infrastructure monitoring, yet existing machine‑ethics models often suffer from one or more key limitations: they either encode rigid, binary rules that cannot accommodate nuanced moral judgments, rely exclusively on data‑driven learning that inherits biases and lacks transparency, or treat ethical concerns in isolation without a unified risk‑centric perspective. Moreover, few approaches systematically integrate expert confidence and stakeholder priorities, leaving designers with little guidance on how to weigh competing harms under uncertainty. Consider a home‑care robot faced with a reluctant patient who refuses medication: should it persist, seek caregiver assistance, or defer entirely? Without a structured mechanism to quantify risks, physical harm from missed doses, autonomy violation through insistence, or loss of patient trust, decisions become ad hoc and opaque. 

ERA appears to be a crucial phase in the Ethical Decision Making (EDM) process for the case of SAI systems and poses several issues.
A major problem is the difficulty of accurately estimating the possible ethical risks without a complete understanding of all aspects of the risk system being studied. In practical scenarios, it is impossible to completely eliminate gaps in ERA, resulting in fuzziness (imprecision, vagueness, incompleteness, etc.) that we need to address and manage.

\paragraph{Contribution}
ERA is inherently a complex and subjective process, largely due to the presence of uncertainty, imprecision, and incomplete or missing data. In many real-world scenarios—especially those involving novel or context-sensitive AI applications—reliable empirical data may be scarce or unavailable. In such cases, it becomes essential to incorporate expert judgment into the assessment process in a structured and traceable manner. To address these challenges and support a flexible, transparent implementation of ERA, we propose a novel framework (ff4ERA) that combines multiple decision-making techniques. Specifically, we leverage fuzzy set theory to capture the inherent vagueness and gradation in ethical evaluations, and apply the FAHP to integrate and weight expert preferences regarding different ethical risk types. This hybrid approach enables the aggregation of both quantitative sensor data and qualitative expert insights, facilitating a comprehensive and interpretable assessment of ethical risks under uncertainty.

We introduce \textbf{ff4ERA}, with two principal contributions:

\begin{enumerate}
  \item A transparent framework for ethical risk assessment: 
    A unified, fuzzy logic–based methodology that quantifies multiple ethical risks producing an interpretable Ethical Risk Score (ERS) to support EDM under a risk‑based governance. Our proposed ff4ERA framework addresses the above mentioned gaps by combining fuzzy logic to model gradations of risk and expert certainty, Mamdani inference for transparent rule evaluation, and FAHP weighting to encode stakeholders priorities.

  \item A comprehensive validation strategy:  
    An integrated local and global sensitivity analysis pipeline—including one‑at‑a‑time perturbations and Sobol variance decomposition—to verify five formal axioms (monotonicity, weight–influence consistency, sub‑evidence dominance, normalization invariance, interaction non‑negativity), ensuring model robustness and transparency.
\end{enumerate}
By quantifying ethical risks in a way that directly supports the EU AI Act’s risk‑based governance (from high‑risk classification to system‑level mitigation), ff4ERA offers both designers and regulators a transparent, data‑driven decision support tool.

\paragraph{Structure}
The remainder of this paper is organized as follows.  Section\ref{lite} reviews computational machine ethics.  
Section \ref{fuzzy} gives some general background about fuzzy logic.
Section\ref{era} details the ff4ERA framework methodology.  Section\ref{case} presents an application of the framework on a concrete case study. Then, Section \ref{dis} discusses the results obtained from applying the framework to the case study of care robot. Finally, we conclude in Section\ref{con} and discuss future works.  

\section{Related Works}
\label{lite}

Machine ethics has been pursued through multiple computational paradigms. Early top‑down (rule‑based) systems encode explicit ethical norms or principles derived from philosophy into logic or case‑based rules \cite{Allen2005, Anderson2006}. For example, Allen \textit{et al.}\ \cite{Allen2005} define “top‑down” methods as translating pre‑existing moral rules (e.g.\ in deontic logic) into a working system . Such logicist or rule‑based frameworks are predictable but rely heavily on formalizing often vague human norms.  

In contrast, bottom‑up approaches let agents learn ethical behavior from data or experience. Recent work treats moral decision‑making as a learning problem: for example, Abel et al.\ argue that an agent’s ethical choices can be modeled as solving a (partially‑observable) Markov decision process (POMDP) in a reinforcement learning framework \cite{Abel2016}. These approaches (often using Deep Learning or RL) can discover complex policies but risk inheriting biases if training data are flawed. Hybrid systems combine both: they use core ethical rules as a scaffold while refining or overriding them through learning. Allen et al.  \cite{Allen2005}\ note that ``both top‑down and bottom‑up approaches embody different aspects of a sophisticated moral sensibility'', and that hybrid combinations can cover shortcomings of either alone. In practice, many machine‑ethics architectures mix rule‑based constraints (e.g.\ deontic logic) with utility‑based reasoning to handle conflicts \cite{Briggs2015, Tolmeijer2020}.  

Fuzzy logic has been proposed as a natural way to handle the uncertainty and gradation inherent in moral judgments. Unlike binary allowed/forbidden rules, fuzzy systems map ethical inputs to continuous degrees of obligation or risk. For instance, Dyoub and Lisi, in \cite{Dyoub2025}, observe that “morality is a fuzzy concept because it lacks clear boundaries and varies according to context,” and they develop a fuzzy rule–based model for ethical decision‑making with formal verification . Their model is based on ERA approach proposed in \cite{DyoubL24}. The ERA system proposed in \cite{DyoubL24} becomes one step (one module) in the current proposed ERA framework, which is more comprehensive. Similarly, Assadi and Inverardi, in \cite{Assadi2024}, explore “functional morality” by encoding human dispositions and contextual ethics into fuzzy membership functions, enabling robots to weigh soft ethical constraints in a continuous manner . These works highlight that fuzzy logic can model the spectrum of moral considerations (e.g.\ risk of harm, privacy violation) and support interpretable rule‑based reasoning. Relatedly, fuzzy Petri nets have been used to represent and verify complex ethical rule sets under uncertainty \cite{Dyoub2025}.  

Decision‑theoretic models (MDPs/POMDPs) and reinforcement learning offer a complementary approach. In this vein, Abel et al.\ formalize ethical choice as a POMDP: an agent must infer a hidden “utility” function representing human values and then optimize actions accordingly \cite{Abel2016}. More recently, Kolker et al. \cite{Kolker2025} \ propose a Multi‑Moral MDP, explicitly modeling multiple conflicting ethical theories (e.g.\ utilitarian vs.\ deontological) as separate objectives under uncertainty . Their approach plans over sequences of actions to satisfy long‑term ethical goals, balancing the trade‑offs between different moral utilities. Such decision‑theoretic frameworks excel at quantifying outcomes and handling stochastic environments, but they require careful design of reward or cost structures that encode moral preferences \cite{Abel2016, Kolker2025}.  

In comparison, rule‑based systems make decisions via symbolic inference (as in “if‑then” ethical rules) \cite{Allen2005}. Both paradigms have been studied: for example, Briggs and Scheutz \cite{Briggs2015} use deontic logic to reject commands that conflict with duties, whereas others use dynamic utility‑maximization. The key distinction is whether ethics are encoded as hard constraints (rules) or as elements of a utility function to be optimized.  

Finally, growing attention has been paid to ethical risk assessment as a complementary framework. Rather than focusing only on moral theory, some recent works treat ethics in AI as managing risk to stakeholders. Dyoub and Lisi \cite{Dyoub2025} emphasize Ethical Risk Assessment (ERA): they argue that AI systems must identify and mitigate risks of harm (physical, privacy, bias, etc.) by selecting actions that minimize fuzzy measures of risk. Douglas et al. \cite{Douglas2024}\ define “ethical risk” in socio‑technical terms—any AI‑related risk that causes stakeholders to fail their ethical responsibilities—and analyze it in terms of stakeholder roles and domination . On the practical side, Murashova et al. \cite{Murashova2025}\ present a methodology for embedding ethical considerations into standard risk assessment processes (e.g.\ using CORAS), showing that a risk‑oriented, multi‑stakeholder approach can operationalize ethics in system design . These risk‑based frameworks complement direct moral reasoning by systematically evaluating potential harms and duty violations of AI behaviors, and have inspired the incorporation of probabilistic risk metrics into machine‑ethics models.  

In summary, the literature spans rule‑based (logicist), learning‑based (decision‑theoretic), and hybrid machine‑ethics models \cite{Allen2005, Anderson2006, Abel2016, Briggs2015, Tolmeijer2020}. Fuzzy logic approaches add a means to represent imprecise moral judgments \cite{Dyoub2025, Assadi2024}. Meanwhile, new frameworks stress ethical risk and safety, combining these techniques to assess and mitigate harm \cite{Douglas2024, Murashova2025}. 

The work presented in this paper aims at establishing ERA as a foundational step toward risk-aware EDM in AI systems. By systematically identifying, quantifying and prioritizing potential ethical risks, such as physical harm, autonomy violations, and trust erosion, the proposed framework enables AI agents to reason about the ethical implications of their actions under uncertainty. This ethical risk centered approach not only supports compliance with emerging regulatory frameworks like the EU AI Act but also promotes transparency, interpretability, and accountability in EDM. 

To the best of our knowledge, this approach to EDM grounded in ERA is novel in the machine ethics literature. Unlike traditional models that rely solely on predefined ethical theories or data-driven learning, our framework systematically quantifies ethical risks under uncertainty to be later integrated into a transparent, interpretable decision-making process. We are currently working on the complete ERA-based ethical decision making system.

\section{Fuzzy Logic and Applications}
\label{fuzzy}
Developed by Lotfi Zadeh\footnote{\url{https://spectrum.ieee.org/lotfi-zadeh}} in the 1960s, fuzzy logic \cite{zadeh1988fuzzy} is based on fuzzy set theory, which is a generalization of the classical set theory. The classical sets are also
called clear sets, as opposed to vague, and similarly classical logic is also
known as Boolean logic or binary. 
A \textit{fuzzy set} is a mathematical construct that allows an element to have a gradual degree of membership within the set, as opposed to the binary inclusion found in classical sets \cite{zimmermann2011fuzzy}. Formally, a fuzzy set \( A \) in a universe of discourse \( X \) is defined by a \textit{membership function} \( \mu_A : X \rightarrow [0, 1] \), where each element \( x \in X \) is assigned a degree of membership \( \mu_A(x) \). This value represents the extent to which \( x \) belongs to the fuzzy set \( A \). Membership functions (MF) can take various shapes, such as triangular, trapezoidal, or Gaussian, depending on the problem domain and the nature of the input data \cite{ross2005fuzzy}.

The concept of MF discussed above allows us to define fuzzy systems in natural language, as the MF couples fuzzy logic with linguistic variables.
Let \( V \) be a variable (e.g., quality of service in a restaurant, tip amount), \( X \) the range of values of the variable, and \( T_V \) a finite or infinite set of fuzzy sets. A \textit{linguistic variable} corresponds to the triplet \( (V, X, T_V) \).

In fuzzy logic, reasoning, also known as \textit{approximate reasoning}, is based on fuzzy rules that are expressed in natural language using linguistic variables such as "HIGH" or "LOW", which we have defined above. A \textit{fuzzy rule} has the form:
\[
\text{If } x \in A \text{ and } y \in B \text{, then } z \in C,
\]
where \( A \), \( B \), and \( C \) are fuzzy sets.
For example: \[
\text{'If (the quality of the food is HIGH), then (tip is HIGH)'.}
\]

Fuzzy logic is particularly effective in systems that must emulate human decision-making. It enables computers and other systems to make decisions based on imprecise or incomplete information, reflecting the way humans process information and make judgments in everyday situations. Fuzzy logic is used in a variety of applications, including consumer electronics (e.g., washing machines, cameras) to industrial control systems (e.g., chemical plant processes, automotive systems), control systems, decision support systems, and pattern recognition \cite{singh2013real,tamir2015fifty}. In healthcare, fuzzy logic can be applied to diagnose conditions, tailor treatments, and optimize resource allocation, ensuring that decisions accommodate the nuances of human health and well-being \cite{thukral2019medical}.

Fuzzy logic offers a flexible framework for handling uncertainties and ambiguities associated with complex decision making processes. Notably, it has been applied for risk assessment and management in many domains. Herein, we highlight some of these applications. 
One of the main applications is in the evaluation of environmental risks, such as pollution levels or the impact of climate change. For instance, fuzzy logic has been used to assess the risk of water pollution by integrating various indicators, such as chemical concentrations, water PH, and temperature, into a single risk index \cite{rea2022risk}.
Another example of application for fuzzy logic is the assessment of risks in work places where data might be vague or incomplete. A fuzzy framework was used for assessing the risk of injury due to machinery, considering hazardous factors such as the skill level of the operator, and the working environment  \cite{tadic2012fuzzy}. This approach allows safety managers to better prioritize risks and implement more effective mitigation strategies. 
Financial risk management is another area in which fuzzy logic was applied. Precise financial risk prediction is very challenging because financial markets are characterized by high levels of uncertainty and volatility. Fuzzy logic helps in modeling such uncertainty, allowing for better decision-making in areas such as portfolio management and credit risk assessment \cite{korol2012fuzzy}.
Fuzzy logic has been also used for assessing and managing the risks associated with project timelines, costs, and resources. Project managers can develop more realistic schedules and budgets, by incorporating fuzzy inputs like the likelihood of delays, cost overruns, and resource availability. In large and complex projects, traditional risk management approaches may fall short due to the high levels of uncertainty involved, fuzzy logic can  offer a valuable solution \cite{MORENOCABEZALI2020106529}.

\section{Proposed Framework for ERA}
\label{era}

\begin{figure}[htb!]
	\centering
	\includegraphics[width=0.45\linewidth]{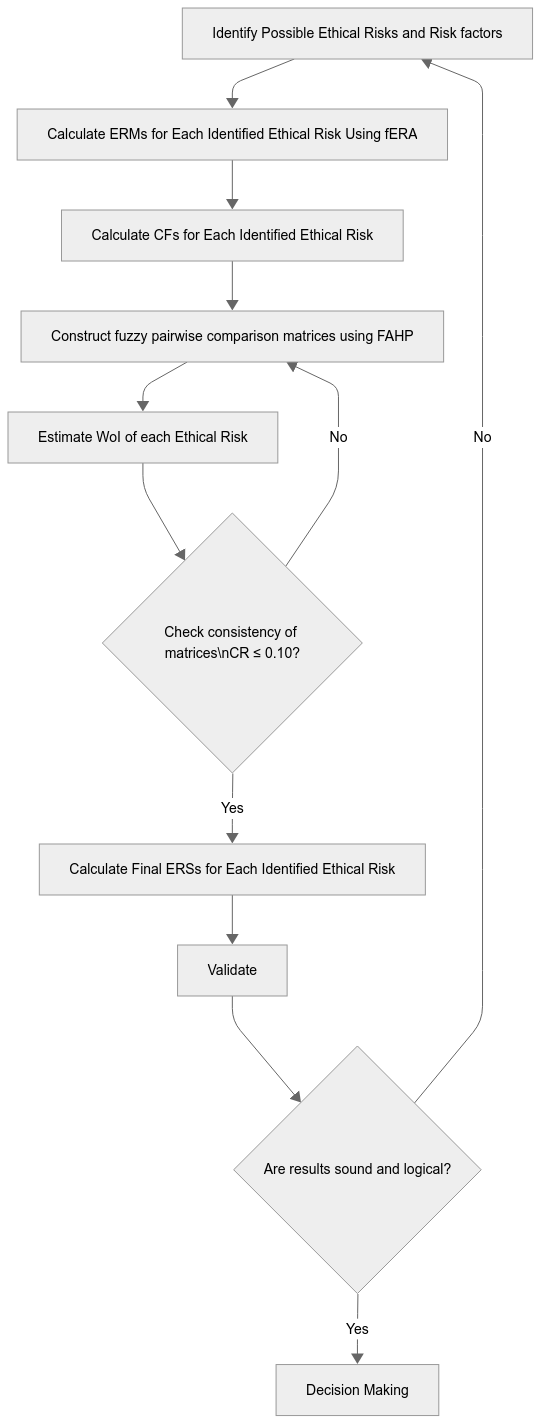}
	\caption{ff4ERA Framework}
	\label{fig:ff4era}
\end{figure}

The proposed framework, shown in Figure \ref{fig:ff4era}, is capable of quantifying qualitative judgements of experts and allows for a step by step analysis of the case at hand in a transparent manner according to the following steps:

\begin{enumerate}
\item Identify possible types of ethical risks and their relevant factors in the case at hand. 

\item For each type of ethical risk, calculate the level/magnitude (ERM) of the Ethical risk using the Fuzzy Ethical Risk Assessment (fERA) system presented in \cite{DyoubL24}. 
\item Assign degrees of belief (CF) to the input factors and to the fuzzy rules with the help of domain experts. These CFs of inputs and rules are used to calculate the CF of the ERM (the output) calculated in the previous step.
\item Calculate weights of importance (WoI) for each type of ethical risk via FAHP. This step involves domain experts.
\item Calculate the Ethical Risk Score (ERS) for each risk type by aggregating the above three values (ERM, CF, and WoI).  This score will tell us how impactful this risk is in the overall ethical decision making context. 
\item Validation: To validate our model, we conduct a comprehensive sensitivity analysis.
\end{enumerate}

\subsection{Identify Ethical Risks and Factors (Step 1)}
\label{identify}
Identify the possible ethical risks in the case at hand. Then, for each type of ethical risk, identify the relevant factors (parameters) that determine its likelihood and potential impact, as these factors are used as inputs to calculate the overall risk level. Furthermore, the corresponding linguistic variables of the ethical risks and their parameters should be defined.

We suggest presenting these ethical risks and their factors in a hierarchical structure. Putting the problem in a hierarchical structure is crucial for providing decision makers with a clear and comprehensive view of the problem.

\subsection{ERM Calculation (Step 2)}
\label{erm}
\begin{figure}[htb!]
	\centering
	\includegraphics[width=0.9\linewidth]{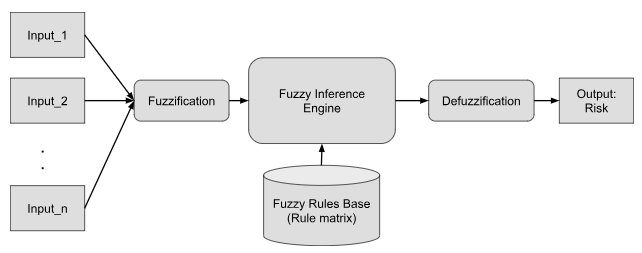}
	\caption{Architecture of our fuzzy system for ERA}
	\label{fig:fuzzymodel}
\end{figure}

In this step, we calculate the ERM for each ethical risk type using fERA.
Figure \ref{fig:fuzzymodel} shows the building blocks of fERA.

The main components of our fERA  system are:
\begin{description}
    \item[Inputs:] These are the factors/parameters relevant for the ethical risk calculation.
	\item[Fuzzification:] In this stage crisp input values are converted into fuzzy sets, , allowing real-world data (e.g., temperature, speed) to be interpreted in a way that accounts for uncertainty or vagueness. This is done using membership functions that map input values to a degree of membership between 0 and 1. For example, in a temperature control system, a crisp input of 75°F might be partially categorized as both “warm” and “hot,” with different membership degrees for each. A \textit{fuzzy set} is a mathematical construct that allows an element to have a gradual degree of membership within the set, as opposed to the binary inclusion found in classical sets \cite{zimmermann2011fuzzy}. Formally, a fuzzy set \( A \) in a universe of discourse \( X \) is defined by a \textit{membership function} \( \mu_A : X \rightarrow [0, 1] \), where each element \( x \in X \) is assigned a degree of membership \( \mu_A(x) \). This value represents the extent to which \( x \) belongs to the fuzzy set \( A \). Membership functions (MF) can take various shapes, such as triangular, trapezoidal, or Gaussian, depending on the problem domain and the nature of the input data \cite{ross2005fuzzy}. For instance, a triangular MF \( \mu_{\mathrm{Tri}}(x; a,b,c) \) is defined as follows:
    
\begin{equation}
\mu_{\mathrm{Tri}}(x; a,b,c) =
\begin{cases}
0, & x \le a,\\[6pt]
\dfrac{x - a}{b - a}, & a < x \le b,\\[8pt]
\dfrac{c - x}{c - b}, & b < x < c,\\[6pt]
0, & x \ge c,
\end{cases}
\label{eq:triMF}
\end{equation}

or, equivalently,

\begin{equation}
\mu_{\mathrm{Tri}}(x; a,b,c)
= \max\!\Bigl(0,\;\min\!\bigl(\tfrac{x - a}{b - a},\,\tfrac{c - x}{c - b}\bigr)\Bigr).
\label{eq:triMF_compact}
\end{equation}

Where \(x\) is the crisp input value (within the universe of discourse).
\(a\) is the lower bound where membership begins (foot of the triangle).
\(b\) is the peak point with full membership (\(\mu=1\)).
\(c\) is the upper bound where membership ends (other foot of the triangle).

The concept of MF discussed above allows us to define fuzzy systems in natural language, as the MF couples fuzzy logic with linguistic variables.
    
	\item[Inference Engine:] The inference engine will consults the  \textit{Fuzzy Rules Base} that contains a set of "if-then" rules that define the system's behavior. A \textit{fuzzy rule} has the form:
\[
\text{If } x \in A \text{ and } y \in B \text{, then } z \in C,
\]
where \( A \), \( B \), and \( C \) are fuzzy sets. These rules describe how fuzzy inputs relate to fuzzy outputs based on expert knowledge. The engine will apply these rules to the fuzzified input to derive fuzzy output sets. It determines which rules are relevant based on the degree of membership of the input values. There are different methods to infer rules, such as the \textit{Mamdani} or \textit{Sugeno} inference methods, which handle how the rules combine to produce a result\footnote{\url{https://it.mathworks.com/help/fuzzy/types-of-fuzzy-inference-systems.html}}. We use the Mamdani method in our case study. Fuzzy rules could be automatically generated from data. In the current implemented version these rules are manually written.
	\item[Defuzzification:] Converting the fuzzy output sets back into crisp values to implement actions or decisions. Common defuzzification methods include \textit{centroid}, \textit{mean of maximum}, and \textit{bisector}, etc\footnote{\url{https://it.mathworks.com/help/fuzzy/defuzzification-methods.html}}. \textit{Centroid} method is the most widely used
methods amongst all the de-fuzzification methods \cite{Lee1990FuzzyLI}. This method provides a center of the area under the curve of the membership function as follows:
\begin{center}
$x_{centroid}= \frac{\sum_{i}\mu (x_i)x_i}{\sum_{i}\mu (x_i)},$
\end{center}
where $x_{centroid}$ is computed using the following formula, where $\mu(x_i)$ is the membership value for point $x_i$ in the universe of discourse.
\item[Output:] The only Output in our fuzzy system is the ethical risk level. 
\end{description}

\subsection{CF Calculation (Step 3)}
\label{cf}
Certainty factors (CFs) play a crucial role in fuzzy‐logic reasoning systems by quantifying the expert’s confidence in each rule or antecedent under uncertainty. While standard fuzzy inference evaluates the degree to which inputs belong to linguistic categories, it does not account for the reliability of the rules themselves or the quality of the underlying data. By assigning a $CF \in [0,1]$ to each fuzzy rule and to each input’s membership degree, we effectively weight the influence of that rule or input on the final conclusion. In practice, this means that even if an antecedent has a high fuzzy‐membership value, a low CF will attenuate its effect in the aggregation phase, reducing the risk of over‐committing to imprecise or incomplete information. Conversely, a high CF amplifies the impact of highly reliable evidence. The combination of fuzzy membership and CF propagation thus yields a more nuanced and robust reasoning process, enabling the system to gracefully degrade its conclusions when information is sparse or uncertain, and to reflect expert trust levels in complex, real‐world decision‐making scenarios.

Based on logical equivalence, logical rules of these two forms: 1) $P_1 \wedge P_2 \wedge ... \wedge P_{j-1} \rightarrow P_j \wedge P_{j+1} \wedge ... \wedge P_k$; 2) $P_1 \vee P_2 \vee ... \vee P_{j-1} \rightarrow P_j \wedge P_{j+1} \wedge ... \wedge P_k$. can be normalized into the following three rule types:
\begin{itemize}
	\item Type 1: $P_1 \wedge P_2 \wedge ... \wedge P_{j-1} \rightarrow P_i$, where $1 < j \geq i\leq k$.	
    \item Type 2: $P_i \rightarrow P_1 \wedge P_2 \wedge ... \wedge P_{j-1} $, where $1\geq i\leq j-1$. This rule can be divided into a set of rules: $P_i \rightarrow P_1$, $P_i \rightarrow P_2$, ..., $P_i \rightarrow P_{j-1}$.
	\item Type 3: $(P_1 \vee P_2 \vee ... \vee P_{j-1})\rightarrow P_j$, where $1 < j \geq i\leq k$.
\end{itemize}

CF is a measure of confidence or belief that quantifies how certain we are about the rule’s conclusion based on the conditions \cite{60794}.
Let $\alpha_i$ denote the degree of truth of antecedent / consequent parts $P_i$ of a rule $r_i$ and $\beta_i$ denotes the degree of confidence of the rule $r_i$. We can obtain the rules with certainty factors as follows: 
\begin{itemize}
	\item Type 1: $R_{i}(\beta_i): P_1 (\alpha_1) \wedge P_2 (\alpha_2)\wedge ... \wedge P_{j-1} (\alpha_{j-1})\rightarrow P_j (\alpha_j)\wedge P_{j+1} (\alpha_{j+1})\wedge ... \wedge P_k (\alpha_k)$
 
	\item Type 2: $R_1(\beta_1) : P_j(\alpha_j) \rightarrow P_1(\alpha_1); 
R_2(\beta_2) : P_j(\alpha_j) \rightarrow P_2(\alpha_2);  \dots ; \\
R_j(\beta_{j-1}) : P_j(\alpha_j) \rightarrow P_{j-1}(\alpha_{j-1}).$

	\item Type 3: $R_{i}(\beta_i): (P_1 (\alpha_1)\vee P_2 (\alpha_2)\vee ... \vee P_{j-1} (\alpha_{j-1}))\rightarrow P_j (\alpha_j)$
\end{itemize}

We use the following rules from \cite{60794}, to calculate the CFs of the calculated ERMs:

\begin{itemize}
	\item Type 1: $R_{i}(\beta_i): P_1 (\alpha_1) \wedge P_2 (\alpha_2)\wedge ... \wedge P_{j-1} (\alpha_{j-1})\rightarrow P_j (\alpha_j)\wedge P_{j+1} (\alpha_{j+1})\wedge ... \wedge P_k (\alpha_k)$ \\
 $\alpha_j = \alpha{j+1} = ... = \alpha_k = min\{\alpha_1, \alpha_2, ... \alpha_{j-1}\}*\beta_i .$
 
	\item Type 2: $R_1(\beta_1) : P_j(\alpha_j) \rightarrow P_1(\alpha_1); 
R_2(\beta_2) : P_j(\alpha_j) \rightarrow P_2(\alpha_2); \dots ;
\\R_j(\beta_{j-1}) : P_j(\alpha_j) \rightarrow P_{j-1}(\alpha_{j-1}).$ \\
$\alpha_1 = \alpha_j * \beta_1 . \alpha_2 = \alpha_j * \beta_2 ... \alpha_{j-1} = \alpha_j * \beta_{j-1} .$

	\item Type 3: $R_{i}(\beta_i): (P_1 (\alpha_1)\vee P_2 (\alpha_2)\vee ... \vee P_{j-1} (\alpha_{j-1}))\rightarrow P_j (\alpha_j)$ \\
 $\alpha_j = max \{\alpha_1 , \alpha_2 , ... , \alpha_{j-1} \} * \beta_i .$
\end{itemize}

\subsection{Calculate WoI via FAHP (Step 4)}
The relative importance of different ethical risks can change dramatically depending on the context in which a decision is made. For example, privacy concerns may loom largest when handling sensitive personal data, whereas fairness and bias issues might take precedence in automated hiring systems. To manage this variability effectively, their weights have to be taken into account in order to represent their relative importance to the overall ethical decision.

Ethical risks weights are determined by FAHP. FAHP is a multi attribute decision making (MADM) technique used to determine weights using fuzzy rules \cite{Demirel2008}. Compared to the conventional AHP
method, which uses crisp values in evaluating the relative importance of each attributes, FAHP uses fuzzy numbers instead of crisp values to ease expert knowledge elicitation. 
 
When evaluating a set of attributes, the technique’s primary aim is to elicit judgments about their relative importance and to translate those judgments into a numerical form that supports easy quantitative analysis (\cite{Demirel2008}).  To assign weights, experts perform pairwise comparisons based on an estimation scheme, which lists the intensity of importance using linguistic terms. Each term corresponds to a triangular fuzzy number (TFN) ${a}_{x} = (L,M,U)$. A decision matrix is formed, on the bases of fuzzy rules, for pair wise comparisons. The values in the decision matrix are dependent on fuzzy membership function. For defining fuzzy rules, triangular fuzzy membership function (TFM) with real numbers is used (defined in (1)).

Table~\ref{tab:weight_scheme} lists TFNs for linguistic variables, as modified and adopted from \cite{varshney2024fuzzy}. It is possible to adopt the scheme that we find suitable for our case.

\begin{table}[ht]
  \centering
  \caption{Weight estimation scheme (linguistic terms $\to$ TFNs)}
  \label{tab:weight_scheme}
  \begin{tabular}{@{}ll@{}}
    \toprule
    \textbf{Level of importance} & \textbf{(TFNs)} \\
    \midrule
    Equal importance    & (1,\,1,\,1) \\
    Moderate importance & (2,\,3,\,4) \\
    Strong importance   & (4,\,5,\,6) \\
    Very strong         & (6,\,7,\,8) \\
    Extreme importance  & (8,\,9,\,10) \\
    \bottomrule
  \end{tabular}
\end{table}

The detailed procedure for determining weights using FAHP is discussed below:

\textbf{Step 1: Identification of criteria and their relative significances} \\[6pt]
In FAHP, criteria are needed to be defined for decision making, which are termed as alternatives and attributes. Let there be $N$ alternatives and $M$ attributes. The weights corresponding to attributes are denoted as $O_m$, where $m = 1, 2, \cdots, M$, and those corresponding to alternatives are denoted as $O_n$, where $n = 1, 2, \cdots, N$.

\textbf{Step 2: Pair-wise decision matrix formulation} \\[6pt]
After defining the alternatives and attributes, a pair-wise decision matrix is formed using TFM function. The elements of the matrix are fuzzy elements taken from Table~\ref{tab:weight_scheme} and are denoted by $a_{m,n}$ and their significance level is decided on the basis of $m^{th}$ attribute's relation with $n^{th}$ alternative. For example, if $m^{th}$ attribute is at "Highest" significance level with respect to $n^{th}$ alternative, then $a_{m,n}$ will be ``$(8,9,10)$'' which is considered from Table~\ref{tab:weight_scheme}. And if there is no difference in the significance level of $m^{th}$ attribute and $n^{th}$ alternative, then $a_{m,n}$ will be considered as 'Identical' i.e., ``$(1,1,1)$". The representation of decision matrix is given in (2).

\begin{equation}\tag{2}
A = \left[ \begin{array}{c|ccccc}
\multicolumn{1}{c}{} & O_1 & O_2 & O_3 & \cdots & O_N \\
\cline{2-6}
O_2 & a_{1,1} & a_{1,2} & a_{1,3} & \cdots & a_{1,N} \\
O_3 & a_{2,1} & a_{2,2} & a_{2,3} & \cdots & a_{2,N} \\
\vdots & \vdots & \vdots & \vdots & \ddots & \vdots \\
O_M & a_{M,1} & a_{M,2} & a_{M,3} & \cdots & a_{M,N}
\end{array} \right]
\end{equation}

Fuzzy element $a_{m,n}$ i.e., TFM function is defined such that $a_{m,n} = a_{n,m}^{-1}$ when $m \ne n$ and $a_{m,n} = 1$ when $m = n$.

Suppose we have $n$ experts in the ethical risk assessment group, then, the elements in the fuzzy pairwise comparison matrix can be modeled as follows aggregating their judgments: 
\[
{a}_{ij} = \frac{1}{n} \otimes ({e}^1_{ij} \oplus {e}^2_{ij} \oplus \cdots \oplus {e}^n_{ij}),\quad
{a}_{ji} = \frac{1}{{a}_{ij}},
\]
where ${a}_{ij}$ is the relative importance by comparing attributes $i,j$, while ${e}^k_{ij}$ is the $k{th}$ expert judgment in TFN format.

\textbf{Step 3: Evaluation of geometric mean} \\[6pt]
The interval arithmetic for TFM function is utilized to evaluate geometric mean ($GM_m$) of the $m^{th}$ alternative which is calculated using (3).

\begin{equation}\tag{3}
GM_m = \left[ \prod_{n=1}^{N} a_{m,n} \right]^{1/N}
\end{equation}

where, $GM_m$ is geometric mean and it shows radical root of $m^{th}$ alternative's in decision matrix.

\textbf{Step 4: Calculation of fuzzy weights} \\[6pt]
For respective attributes, relative fuzzy weights ($FO_m$) are calculated as

\begin{equation}\tag{4}
FO_m = \frac{GM_m}{\sum\limits_{m=1}^{M} GM_m}
\end{equation}

\textbf{Step 5: Calculation of best non-fuzzy performance value as weights} \\[6pt]
The calculation of best non-fuzzy performance (BNFP) value as weights is done as

\begin{equation}\tag{5}
W_m = \frac{\text{FO(L)}_m + \text{FO(M)}_m + \text{FO(U)}_m}{3}
\end{equation}

where \(\text{FO(L)}_m\), \(\text{FO(M)}_m\) and \(\text{FO(U)}_m\) represent the lower, middle and upper fuzzy values, respectively, to calculate BNFP value based on fuzzy membership function.

\textbf{Step 6: Consistency Ratio in FAHP} \\[6pt]
In FAHP, the \emph{Consistency Ratio} (CR) is a measure of how logically consistent our pairwise comparisons are.  When we compare items two‐at‐a‐time (say, criteria or alternatives) on a numerical “importance” scale (1=equal up to 9=extreme preference), we build an $n\times n$ reciprocal matrix $A$ (so that $a_{ij}=1/a_{ji}$).  If our judgments were perfectly self‐consistent, we’d have
\[
a_{ik} \;=\; a_{ij}\,\times\,a_{jk}
\]
for every triple $(i,j,k)$, and the largest eigenvalue $\lambda_{\max}$ of $A$ would equal~$n$.

In practice, judgments are rarely perfect, so the maximum weight value of an n.by-n comparison matrix $\lambda_{\max}>n$. Saaty in \cite{saaty1990make} shows that the “degree of inconsistency” can be captured by how far $\lambda_{\max}$ exceeds $n$, namely
\[
\text{Consistency Index (CI)}
\;=\;
\frac{\lambda_{\max}-n}{\,n-1\,}.
\]
This normalizes the raw deviation $(\lambda_{\max}-n)$ by the matrix size $(n-1)$, giving an average “inconsistency per comparison.”

\[
\lambda_{\max} = \frac{1}{n} \sum_{j=1}^n \frac{\sum_{k=1}^n a_{jk} w_k}{w_j}
\]

CI by itself has no scale, we need to compare it to what we’d get from purely random judgments.  For each $n$, statisticians have estimated the average CI of many random reciprocal matrices (the “Random Index,” RI).  Saaty’s classic table \cite{saaty2005theory} is:

\begin{table}[h]
  \centering
  \caption{Random Index (RI) values for different matrix sizes}
  \label{tab:ri}
  \begin{tabular}{@{}c c c c c c c c c c c@{}}
    \toprule
    $n$    & 1 & 2 & 3   & 4    & 5    & 6    & 7    & 8    & 9    & 10   \\
    \midrule
    RI     & 0 & 0 & 0.58 & 0.90 & 1.12 & 1.24 & 1.32 & 1.41 & 1.45 & 1.49 \\
    \bottomrule
  \end{tabular}
\end{table}

Finally, the \emph{Consistency Ratio} is
\[
\mathrm{CR}
\;=\;
\frac{\mathrm{CI}}{\mathrm{RI}}.
\]
If $\mathrm{CR}<0.10$ (10\%), judgments are acceptably consistent.  If $\mathrm{CR}\ge0.10$, we should revisit the most “offending” comparisons and re‐examine our judgments.


\subsection{Calculate the Final ERS for each Ethical Risk (step 5)}

The ERS represents the criticality degree of each possible ethical risk. It is calculated using the following formula:

\begin{equation}
ERS = ERM * CF * WoI
\label{eq:ers_formula}
\end{equation}

\subsection{Validation Testing (Step 6)}
\label{sens}
When a new methodology is developed, it requires a careful test to ensure its soundness. It may be especially important and desirable when subjective elements are involved in the methodology generated.
In our framework, we produce a single numerical score for each ethical risk in a given scenario, but that score alone does not tell us which inputs have the greatest influence nor how robust it is to small changes.  That’s where \textit{Sensitivity Analysis} (SA) \cite{contini2000sensitivity} comes in. In models involving many input variables, SA is an essential ingredient for model building and quality assurance. By systematically varying each input factor, one at a time or in coordinated clusters, SA reveals which parameters have the greatest effect on the final risk estimate.  Decision makers can then focus their attention on those “weakest links” in the model: the assumptions or measurements that, if tweaked even slightly, produce the largest swings in ethical risk.  Based on this insight, design teams can prioritize data collection efforts, tighten controls around volatile variables, or redesign processes to reduce the system’s overall vulnerability.

We applied SA to understand how small changes in inputs affect the final risk score. Specifically, minor variations in the model parameters or changes in the degrees of belief assigned to linguistic variables used to describe the parameters by experts. By understanding how inputs affect output, SA can inform decision-making processes. SA  can help validate the model by ensuring that it behaves as expected when input parameters are varied. 

If the proposed framework methodology is sound and its inference reasoning is logical then SA must satisfy the following criteria:

\begin{enumerate}
  \item \textit{Monotonicity:}  
    A small increase (decrease) in any input produces a corresponding relative increase (decrease) in the ERS.
  
  \item \textit{Weight–Influence Consistency:}  
    Equal variations in different inputs produce ERS changes proportional to their FAHP weights.

  \item \textit{Sub‐evidence Dominance:}  
    The combined influence of a set of factors always exceeds that of any strict subset.

  \item \textit{Normalization Invariance:}  
    Uniformly scaling all FAHP weights (and re‐normalizing) leaves the relative sensitivities of ERSs unchanged.

  \item \textit{Interaction Non‐negativity:}  
    For any two inputs \(i,j\), the cross‐partial derivative
    \[
      \frac{\partial^2 \mathrm{ERS}}{\partial b_i\,\partial b_j} \;\ge\; 0
    \]
    i.e.\ increasing one factor never reduces the marginal impact of another.
\end{enumerate}

\section{Application of the Framework to a Concrete Case Study}
\label{case}

We will illustrate our framework step by step using a concrete “patient‐dilemma” scenario in a home care setting, where a personal care assistant robot must decide whether to insist on a medical or wellness-related intervention despite a reluctant patient. In this case study:

A care robot supports an elderly or chronically ill patient in their own home. It helps monitor vital signs, encourage medication adherence, and assist with physical wellness tasks (e.g. hydration, walking, alerting caregivers). An ethical dilemma arises when the robot approaches its adult patient to give her her medicine in time and the patient rejects to take it. Should the care robot try again to change the patient’s mind or accept the patient’s decision as final?

We aim to demonstrate how our framework: i) Identifies possible ethical risks involved in this scenario and their relevant parameters. The values of these parameters are collected through subjective observations and sensor inputs. ii) Aggregates these elements to yield risk-level values for the identified ethical risk types. iii) Computes priority weights for different ethical factors using FAHP. iv) Incorporates belief degrees to reflect confidence in input and in fuzzy rules. v) Calculates ERSs to guide intervention priorities. vi) And validates model behavior through structured robustness testing.

The care robot uses the calculated ERSs to inform decisions such as: (i) whether to repeat a recommendation, (ii) alert a remote caregiver, (iii) defer the action, (iv) or override patient resistance only when ethically justifiable.
The framework also supports "what-if" analysis, helping designers and ethicists understand which patient states contribute most to ethical risks and which inputs require more precise sensing or interpretation.
This example demonstrates the full methodology in a concrete scenario. It provides a traceable, explainable path from uncertain inputs to risk-informed ethical decision making, ready to support both autonomous behavior and human oversight.

\subsection{Identify possible ethical risks and their relative factors:}
\begin{figure}[htb!]
	\centering
	\includegraphics[width=0.5\linewidth]{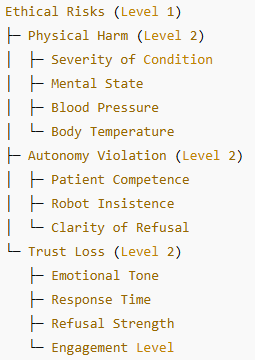}
	\caption{Patient Dilemma Problem model}
	\label{fig:risks}
\end{figure}

The identified ethical risks and risk factors are presented in Figure\ref{fig:risks}. Level 1 defines the types of ethical risks we care about.
Level 2 lists the input factors that feed into each risk’s fuzzy‐logic calculation.
This structural representation helps decision makers to see at a glance both the big picture (which risk types exist) and the detailed drivers (which specific measurements influence each risk).

\subsection{Calculate the ERMs for the Identified Ethical Risks:}

We use fERA to calculate the risk levels (ERMs) for the three identified ethical risks: physical harm (PH), violation of autonomy (AV), and loss of trust (TL).
fERA maps quantitative sensor and behavioral inputs (rated 1–10) into a risk percentage (0\,\%–100\,\%).

\paragraph{Fuzzification}
All input variables $x\in[1,10]$ are fuzzified using TMF (defined in equation\ref{eq:triMF}).
We also use the TMF for output risk $y\in[0,100]$ with three linguistic values: Low, Medium, and High:
\[
\text{Low}=\triMF{0}{0}{50},\quad
\text{Med}=\triMF{25}{50}{75},\quad
\text{High}=\triMF{50}{100}{100}.
\]

\paragraph{Input Variables and Membership Functions}
Tables \ref{ph}, \ref{av}, and \ref{tl} shows the MFs for input parameters of the three possible ethical risks of this case study.
\begin{table}[!hb]
\centering
\caption{Physical Harm Input MFs}
\label{ph}
\begin{tabular}{lccc}
\toprule
Variable & Low & Med & High \\
\midrule
Severity         & $(1,1,5)$ & $(3,5,7)$ & $(5,10,10)$ \\
Mental state     & $1,1,5)$ & $(3,5,7)$ & $(5,10,10)$ \\
Blood pressure   & $(1,1,4)$ & $(3,5,7)$ & $(6,10,10)$ \\
Body temperature & $(1,1,4)$ & $(3,5,7)$ & $(6,10,10)$ \\
\bottomrule
\end{tabular}
\end{table}

\begin{table}[!hb]
\centering
\caption{Autonomy Violation Input MFs}
\label{av}
\begin{tabular}{lccc}
\toprule
Variable & Low & Med & High \\
\midrule
Competence           & $(1,1,5)$ & $(3,5,7)$ & $(5,10,10)$ \\
Robot insistence     & $(1,1,5)$ & $(3,5,7)$ & $(5,10,10)$ \\
Clarity of refusal   & $(1,1,5)$ & $(3,5,7)$ & $(5,10,10)$ \\
\bottomrule
\end{tabular}
\end{table}

\begin{table}[!hb]
\centering
\caption{Trust Loss Input MFs}
\label{tl}
\begin{tabular}{lccc}
\toprule
Variable           & Low & Med & High \\
\midrule
Emotional tone     & $(1,1,5)$ & $(3,5,7)$ & $(5,10,10)$ \\
Response time      & $(1,1,5)$ & $(3,5,7)$ & $(5,10,10)$ \\
Refusal strength   & $(1,1,5)$ & $(3,5,7)$ & $(5,10,10)$ \\
Engagement level   & $(1,1,5)$ & $(3,5,7)$ & $(5,10,10)$ \\
\bottomrule
\end{tabular}
\end{table}

\paragraph{Rule Bases}
We employ Mamdani inference with the minimum–maximum operators.  In this case study, we used the following fuzzy inference rules.

Rules for Physical Harm (PH):
\begin{enumerate}
  \item Rule PH-1: IF Severity is High OR Blood pressure is High OR Temperature is High THEN PH is High
  \item Rule PH-2: IF Severity is Medium AND Mental state is Medium THEN PH is Medium
  \item Rule PH-3: IF Severity is Low AND Mental state is High THEN PH is Low
\end{enumerate}

Rules for Autonomy Violation (AV):
\begin{enumerate}
  \item Rule AV-1: IF Competence is High AND Robot insistence is High THEN AV is High
  \item Rule AV-2: IF Robot insistence is medium AND Clarity is unclear THEN AV is Low
  \item Rule AV-3: IF Competence is Medium AND Robot insistence is Low THEN AV is Low
  \item Rule AV-4: IF Competence is Low OR Clarity is unclear THEN AV is Low
   \item Rule AV-5: IF Competence is Medium AND Robot insistence is Medium THEN AV is Medium
\end{enumerate}

Rules for Trust Loss (TL):
\begin{enumerate}
  \item Rule TL-1: IF Emotional tone is Frustrated OR Response time is Long THEN TL is High
  \item Rule TL-2: IF Refusal strength is Moderate AND Engagement is Medium THEN TL is Medium
  \item Rule TL-3: IF Emotional tone is Calm AND Engagement is High THEN TL is Low
\end{enumerate}

\paragraph{Input Values and Fuzzification} Table \ref{crisp} shows example crisp inputs together with the fuzzification degrees.
\begin{table}[h]
  \centering
  \caption{Fuzzification of Input Variables}
  \label{crisp}
  \begin{tabular}{lcccc}
    \toprule
    \textbf{Input Variable}          & \textbf{Crisp Value} & \textbf{Low $\mu$} & \textbf{Med $\mu$} & \textbf{High $\mu$} \\
    \midrule
    \multicolumn{5}{l}{\textit{Physical Harm}} \\
    Severity                 & 8  & 0.00 & 0.15 & 0.60 \\
    Mental state             & 6  & 0.00 & 0.50 & 0.05 \\
    Blood pressure           & 7  & 0.00 & 1.00 & 0.25 \\
    Body temperature         & 9  & 0.00 & 0.00 & 0.75 \\
    \midrule
    \multicolumn{5}{l}{\textit{Violation of Autonomy}} \\
    Competence level         & 4  & 0.25 & 0.50 & 0.00 \\
    Robot insistence level   & 7  & 0.00 & 0.00 & 0.40 \\
    Clarity of refusal       & 3  & 0.50 & 0.00 & 0.00 \\
    \midrule
    \multicolumn{5}{l}{\textit{Loss of Trust}} \\
    Emotional tone           & 2  & 0.75 & 0.00 & 0.00 \\
    Response time            & 8  & 0.00 & 0.00 & 0.60 \\
    Refusal strength         & 5  & 0.00 & 1.00 & 0.00 \\
    Engagement level         & 6  & 0.00 & 0.50 & 0.20 \\
    \bottomrule
  \end{tabular}
\end{table}

\paragraph{Inference and Aggregation}
After evaluating the firing strength of each rule and aggregating by maximum, we obtain:
\[
\begin{aligned}
\mu_{\mathrm{PH}} &= [0,\;0.15,\;0.75]\quad (\mathrm{Low},\;\mathrm{Med},\;\mathrm{High}),\\
\mu_{\mathrm{AV}} &= [0,\;0.25,\;0.50],\\
\mu_{\mathrm{TL}} &= [0.05,\;0.50,\;0.75].
\end{aligned}
\]

\paragraph{Defuzzification} 
Each aggregated MF $\mu_{\mathrm{agg}}(y)$ is defuzzified using the centroid method:
\[
y^* = \frac{\int_{0}^{100} y\,\mu_{\mathrm{agg}}(y)\,\mathrm{d}y}{\int_{0}^{100}\mu_{\mathrm{agg}}(y)\,\mathrm{d}y}.
\]
The resulting ethical risks levels/magnitudes (ERMs) are shown in Table \ref{level}.
\begin{table}[h]
\centering
\caption{Defuzzified Risk Levels}
\label{level}
\begin{tabular}{lc}
\toprule
Ethical Risk            & ERM (\%) \\
\midrule
Physical Harm          & 78          \\
Autonomy Violation     & 25          \\
Trust Loss             & 65          \\
\bottomrule
\end{tabular}
\end{table}

\subsection{Calculate the CFs for the Identified Ethical Risks:}

To model the influence of rule confidence and antecedent strength on the certainty of ethical risk assessments, we employ CF propagation approach. We consider the three types of fuzzy rules mentioned in Section \ref{era}.

Below, we provide example CF calculations using fuzzified input values obtained in the previous step.

\textbf{Physical Harm Risk CF (Rule PH-1: Type 3 Rule):}

\begin{center}
\textit{IF Severity is High OR Blood pressure is High OR Temperature is High THEN PH is High}, \quad $\beta_{PH2} = 0.8$
\end{center}

Using the following belief degrees:
\[
\alpha_{\text{Severity=High}} = 0.62, \quad \alpha_{\text{BP=High}} = 0.34, \quad
\alpha_{\text{Temp=High}} = 0.79
\]
\[
\alpha_{\text{PH=High}} = \max(0.62, 0.34, 0.79) \cdot 0.8 = 0.79 \cdot 0.8 = 0.632
\]

\textbf{Autonomy Violation Risk CF (Rule AV-4: Type 3 Rule):}

\begin{center}
\textit{IF Competence is Low OR Clarity is unclear THEN AV is Low}, \quad $\beta_{AV1} = 0.9$
\end{center}

Belief degrees:
\[
\alpha_{\text{Competence=Low}} = 0.45, \quad \alpha_{\text{Clarity=Unclear}} = 0.72
\]
\[
\alpha_{\text{AV=Low}} = \max(0.72, 0.45) \cdot 0.9 = 0.72 \cdot 0.9 = 0.648
\]

\textbf{Trust Loss Risk (Rule TL-1: Type 3 Rule):}

\begin{center}
\textit{IF Emotional Tone is Frustrated OR Response Time is Long THEN Trust Loss is High}, \quad $\beta_{TL1} = 0.7$
\end{center}

Belief degrees:
\[
\alpha_{\text{Emotional=Frustrated}} = 0.00, \quad \alpha_{\text{Response=Long}} = 0.75
\]
\[
\alpha_{\text{TL=High}} = \max(0.00, 0.75) \cdot 0.7 = 0.75 \cdot 0.7 = 0.525
\]

The resulting certainty factors for each risk are:
\[
\alpha_{\text{PH=High}} = 0.632, \quad
\alpha_{\text{AV=Low}} = 0.648, \quad
\alpha_{\text{TL=High}} = 0.525
\]
These values indicate the degree of certainty with which each ethical risk level is inferred from the given fuzzy rules and input conditions.

\subsection{Calculate the WoI for the Identified Ethical Risks Using FAHP:}

To derive the relative importance of the three ethical risks: Physical Harm (PH), Autonomy Violation (AV), and Trust Loss (TL), we apply the FAHP as follows.

\paragraph{Step 1: Fuzzy Pairwise Comparison Matrix}
Experts express pairwise comparisons using triangular fuzzy numbers (TFNs) taken from Tabel \ref{tab:weight_scheme}. For simplicity, we assumed to have only one expert.

The pairwise matrix \(\tilde{A} = [\tilde a_{ij}]\) is:
\[
  \tilde{A} =
  \begin{bmatrix}
    (1,1,1)         & (2,3,4)           & (4,5,6)         \\
    (1/4,1/3,1/2)   & (1,1,1)           & (2,3,4)         \\
    (1/6,1/5,1/4)   & (1/4,1/3,1/2)     & (1,1,1)
  \end{bmatrix},
\]
where rows/columns correspond to \(\{\,\text{PH},\text{AV},\text{TL}\}\).

\paragraph{Step 2: Fuzzy Geometric Means}
For each criterion \(i\):
\[
  \tilde g_i
  = \Bigl(\prod_{j=1}^3 \tilde a_{ij}\Bigr)^{1/3}.
\]
Thus
\[
  \tilde g_1
  = \bigl((1,1,1)\cdot(2,3,4)\cdot(4,5,6)\bigr)^{1/3}
  = (8,15,24)^{1/3}
  \approx (2.00,\,2.47,\,2.88),
\]
\[
  \tilde g_2
  = \bigl((1/4,1/3,1/2)\cdot(1,1,1)\cdot(2,3,4)\bigr)^{1/3}
  \approx (0.79,\,1.15,\,1.58),
\]
\[
  \tilde g_3
  = \bigl((1/6,1/5,1/4)\cdot(1/4,1/3,1/2)\cdot(1,1,1)\bigr)^{1/3}
  \approx (0.40,\,0.57,\,0.79).
\]

\paragraph{Step 3: Normalization of Fuzzy Weights}\mbox{}\\

Sum: 
\(\sum \tilde g_i \approx (3.19,\,4.19,\,5.25)\).  Then
\[
  \tilde w_i
  = \frac{\tilde g_i}{\sum_{k=1}^3 \tilde g_k},
\]
giving
\[
  \tilde w_1 \approx (0.38,\,0.59,\,0.90),\quad
  \tilde w_2 \approx (0.15,\,0.27,\,0.50),\quad
  \tilde w_3 \approx (0.08,\,0.14,\,0.25).
\]

\paragraph{Step 4 and 5: Defuzzification}
Using the centroid formula \(w_i=(l+m+u)/3\):
\[
  w_1 = \frac{0.38+0.59+0.90}{3} = 0.623,
  w_2 = \frac{0.15+0.27+0.50}{3} = 0.307,
  w_3 = \frac{0.08+0.14+0.25}{3} = 0.157.
\]
After normalization \( \sum w_i = 1.087\):
\[
  w_1 = 0.623/1.087 = 0.573,\quad
  w_2 = 0.307/1.087 = 0.282,\quad
  w_3 = 0.157/1.087 = 0.145.
\]

Final weights are shown in Table \ref{fahp}.
\begin{table}[h]
  \centering
  \caption{FAHP Weights for Ethical Risks}
  \label{fahp}
  \begin{tabular}{lcc}
    \toprule
    Ethical Risk       & TFN Weight               & Defuzzified Weight \\
    \midrule
    Physical Harm      & \((0.38,0.59,0.90)\)      & 0.573 \\
    Autonomy Violation & \((0.15,0.27,0.50)\)      & 0.282 \\
    Trust Loss         & \((0.08,0.14,0.25)\)      & 0.145 \\
    \bottomrule
  \end{tabular}
\end{table}
\paragraph{Step 6: Consistency Ratio (CR) Calculation for the FAHP Comparison Matrix}
\mbox{} \\
To ensure the reliability of expert judgments in the pairwise comparison matrix for FAHP, we compute the Consistency Ratio (CR) using Saaty's method.

From Section 4.4, the fuzzy matrix is defuzzified using the middle values of triangular fuzzy numbers, yielding the crisp reciprocal matrix:
\[
A = 
\begin{bmatrix}
1 & 3 & 5 \\
\frac{1}{3} & 1 & 3 \\
\frac{1}{5} & \frac{1}{3} & 1
\end{bmatrix}
\]

The normalized weight vector (from FAHP) is:
\[
w = 
\begin{bmatrix}
0.573 \\
0.282 \\
0.145
\end{bmatrix}
\]

Calculate $\lambda_{\max}$:

We compute the weighted sum vector \( A \cdot w \):
\[
A \cdot w = 
\begin{bmatrix}
1 \cdot 0.573 + 3 \cdot 0.282 + 5 \cdot 0.145 \\
\frac{1}{3} \cdot 0.573 + 1 \cdot 0.282 + 3 \cdot 0.145 \\
\frac{1}{5} \cdot 0.573 + \frac{1}{3} \cdot 0.282 + 1 \cdot 0.145
\end{bmatrix}
=
\begin{bmatrix}
2.274 \\
1.047 \\
0.539
\end{bmatrix}
\]

Then, we divide each element by the corresponding weight:
\[
\frac{A \cdot w}{w} = 
\begin{bmatrix}
\frac{2.274}{0.573} \\
\frac{1.047}{0.282} \\
\frac{0.539}{0.145}
\end{bmatrix}
=
\begin{bmatrix}
3.97 \\
3.71 \\
3.72
\end{bmatrix}
\]

Thus, the principal eigenvalue is:
\[
\lambda_{\max} = \frac{3.97 + 3.71 + 3.72}{3} = 3.80
\]

\[
CI = \frac{\lambda_{\max} - n}{n - 1} = \frac{3.80 - 3}{2} = 0.40
\]

For $n = 3$, the Random Index is:
\[
RI = 0.58
\]

\[
CR = \frac{CI}{RI} = \frac{0.40}{0.58} \approx 0.69
\]

Since $CR = 0.69 > 0.10$, the matrix is considered inconsistent. This indicates that the expert judgments in the FAHP matrix may need revision, especially in comparisons involving Physical Harm and Trust Loss.


\subsection{Calculate the final ERSs for the Identified Ethical Risks:}

Using the formula \ref{eq:ers_formula}, and the values obtained previously:

\begin{itemize}
  \item Weights: \(w_{\mathrm{PH}}=0.573\), \(w_{\mathrm{AV}}=0.282\), \(w_{\mathrm{TL}}=0.145\).
  \item Certainty factors: \(\mathrm{CF}_{\mathrm{PH}}=0.632\), \(\mathrm{CF}_{\mathrm{AV}}=0.648\), \(\mathrm{CF}_{\mathrm{TL}}=0.525\).
  \item Ethical risks levels (\%): \(\text{RL}_{\mathrm{PH}}=78\), \(\text{RL}_{\mathrm{AV}}=25\), \(\text{RL}_{\mathrm{TL}}=65\).
\end{itemize}

We obtain the following ERSs of the three ethical risks:
\[
  \mathrm{ERS}_{\mathrm{PH}} =\; 28.25, \quad
  \mathrm{ERS}_{\mathrm{AV}} =\; 4.57, \quad
  \mathrm{ERS}_{\mathrm{TL}} =\; 4.95.
\]

These ERS values indicate that, under the chosen belief confidences and risk levels, 'Physical Harm' emerges as the dominant concern, far outstripping both 'Violation of Autonomy' and 'Trust Loss'. This suggests that, given the current sensor readings and expert certainties, mitigating physical harm should be the highest priority in any subsequent decision or intervention.

\subsection{Validate the Model Behavior Through Sensitivity Analysis:}

We perform a local sensitivity analysis by perturbating input variables one at a time, and global ((Variance‑based) ) sensitivity analysis using Sobol indices.

\subsubsection{Perturbation of Lower‐Level Inputs}

\begin{figure}[h]
  \centering
  \includegraphics[width=0.7\linewidth]{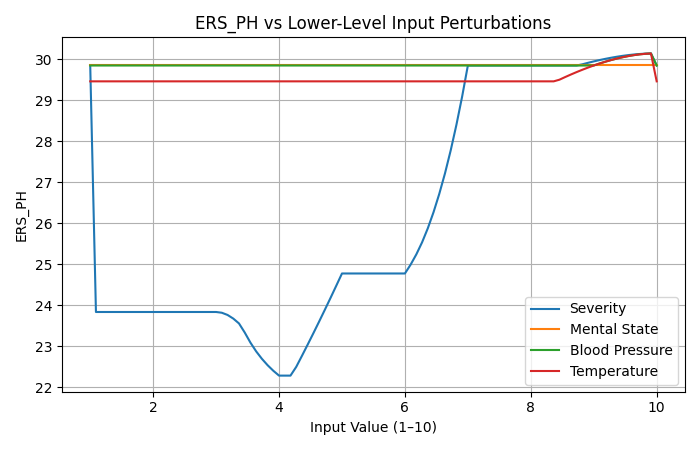}
  \caption{Local sensitivity of ERS$_{\rm PH}$ to each lower‐level input. Severity (blue) shows non‐monotonic behavior due to rule‐dominance shifts; other factors respond monotonically.}
  \label{fig:ers_all_inputs}
\end{figure}

To assess the local sensitivity of the Physical-Harm ERS to its four lower‐level inputs, we performed the following procedure:

\begin{enumerate}
  \item \textbf{Baseline Setup:}  
    \begin{itemize}
      \item Input values: Severity = 8, Mental State = 6, Blood Pressure = 7, Body Temperature = 9.
      \item Fixed parameters: CF$_{\rm PH}=0.632$, $w_{\rm PH}=0.573$.
      \item Membership functions (Triangular):
        \[
          \mu_{\text{Low}}=(1,1,5),\quad
          \mu_{\text{Med}}=(3,5,7),\quad
          \mu_{\text{High}}=(5,10,10).
        \]
    \end{itemize}

  \item \textbf{One‐at‐a‐Time Perturbation:}  
    For each input factor \(x\in\{\text{Severity},\text{Mental},\text{BP},\text{Temp}\}\):
    \begin{enumerate}
      \item Vary \(x\) uniformly from 1 to 10 in 100 steps.
      \item Hold the other three factors at their baseline values.
      \item \emph{Fuzzification:} Compute membership degrees \(\mu_{\!\text{Low,Med,High}}(x)\).
      \item \emph{Inference:}  
        Evaluate the three Mamdani rules for PH:
        \[
        \begin{aligned}
          R_1 &: \max\{\mu_{\text{Sev,High}},\mu_{\text{BP,High}},\mu_{\text{Temp,High}}\}\;\Rightarrow\;\text{PH is High},\\
          R_2 &: \min\{\mu_{\text{Sev,Med}},\mu_{\text{Ment,Med}}\}\;\Rightarrow\;\text{PH is Medium},\\
          R_3 &: \min\{\mu_{\text{Sev,Low}},\mu_{\text{Ment,High}}\}\;\Rightarrow\;\text{PH is Low}.
        \end{aligned}
        \]
      \item \emph{Aggregation \& Defuzzification:}  
        Form the output fuzzy set over \(y\in[0,100]\) with Tri\,\((0,0,50)\), Tri\,(25,50,75), Tri\,(50,100,100) MFs, then compute
        \[
          \mathrm{ERM}_{\rm PH}
          \;=\;\frac{\int_0^{100} y\,\mu_{\rm agg}(y)\,dy}{\int_0^{100} \mu_{\rm agg}(y)\,dy}.
        \]
      \item \emph{ERS Computation:}  
        \(\displaystyle\mathrm{ERS}_{\rm PH}
          = w_{\rm PH}\,\cdot\,\mathrm{CF}_{\rm PH}\,\cdot\mathrm{ERM}_{\rm PH}.\)
    \end{enumerate}

  \item \textbf{Results:}  
    Figure~\ref{fig:ers_all_inputs} shows ERS$_{\rm PH}$ as a function of each input:
    \begin{itemize}
      \item Severity (blue): exhibits a “dip–then–rise” non‐monotonicity.
      \item Mental State (orange): nearly flat, minimal influence.
      \item Blood Pressure (green) and Temperature (red): monotonic increase.
    \end{itemize}
\end{enumerate}

\paragraph{Non‐Monotonic Behavior of Severity}

The “dip–then–rise” observed when varying Severity alone:

\begin{enumerate}
  \item Low‐Severity Region (1–3):  
    Only the “Low” rule contributes (weakly, since \(\mu_{\rm Ment,High}\approx0.05\)), yielding a moderate ERS.
  \item Medium Region (3–7): 
    The “Medium” rule dominates with firing strength \(\min\{\mu_{\rm Sev,Med},\mu_{\rm Ment,Med}\}\approx0.50\), lower than the eventual High‐rule strength, causing ERS to dip.
  \item High‐Severity Region (7–10):  
    The “High” rule takes over (\(\mu_{\rm Sev,High}\) rises to 1.0), pushing ERS back up to its peak.
\end{enumerate}

This behavior is an expected consequence of overlapping triangular MFs and rule certainties.  To enforce strict monotonicity, one may narrow the Medium MF, shift the High MF leftward, or reduce the Medium‐rule certainty so that the High rule never becomes undercut.

\subsubsection{Perturbation of Certainty Factors}

In this step we assess how uncertainties in the fuzzy‐rule certainty factors (CFs) and in the antecedent degrees of belief propagate to the final Ethical Risk Score ERS$_{\rm PH}$.

\paragraph{Perturbation of Rule CF (\(\beta_1\))}

We first vary the certainty factor \(\beta_1\) of the 'High' rule  
\[
R_1:\;\bigl(\text{Sev is High}\lor\text{BP is High}\lor\text{Temp is High}\bigr)
\;\Longrightarrow\;\text{PH is High}
\]
from 0 to 1 in 50 uniform steps.  All other inputs, antecedent CFs, and the FAHP weight \(w_{\rm PH}=0.573\) remain at their baseline values.  The defuzzified risk level \(\mathrm{RL}_{\rm PH}=78\%\) is fixed.

For each \(\beta_1\), the ERS is computed as
\[
\mathrm{ERS}_{\rm PH}
= w_{\rm PH}\;\cdot\;\beta_{1}\;\cdot\;\mathrm{RL}_{\rm PH}.
\]

\begin{figure}[h]
  \centering
  \includegraphics[width=0.6\linewidth]{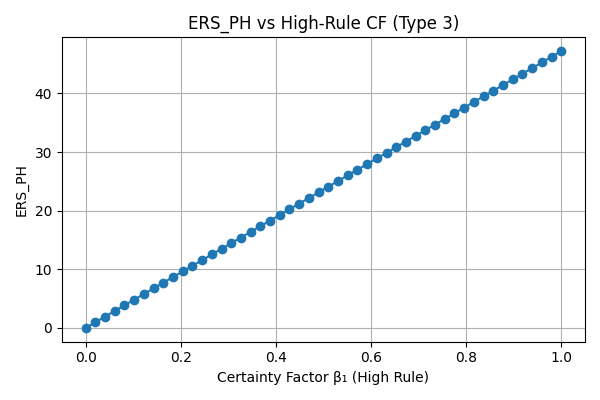}
  \caption{ERS$_{\rm PH}$ as a function of the 'High'‐rule CF \(\beta_1\).  The linear trend confirms \emph{monotonicity} in rule certainty (Axiom1).}
  \label{fig:ers_ph_vs_beta1}
\end{figure}

\begin{table}[h]
  \centering
  \caption{Sample values of ERS$_{\rm PH}$ vs.\ \(\beta_1\).}
  \begin{tabular}{cc}
    \toprule
    \(\beta_1\) & ERS$_{\rm PH}$ \\
    \midrule
     0.00 &  0.00 \\
     0.25 & 11.20 \\
     0.50 & 22.40 \\
     0.75 & 33.60 \\
     1.00 & 44.80 \\
    \bottomrule
  \end{tabular}
\end{table}

\paragraph{Perturbation of Antecedent Degrees of Belief}

Now, we examine the effect of uncertainty in each antecedent’s degree of belief \(\alpha_i\) for the same 'High' rule.  Let the baseline antecedent beliefs be
\[
\alpha_{\text{Sev,High}}=0.62,\quad
\alpha_{\text{BP,High}}=0.25,\quad
\alpha_{\text{Temp,High}}=0.75,
\]
and rule CF \(\beta_1=0.8\).  We vary one antecedent \(\alpha_i\) from 0 up to its baseline, in 50 steps, while holding the other two constant.

Under a Type3 disjunctive rule, the rule’s output CF is
\[
\alpha_{\rm cons}
= \max\{\alpha_i,\;\alpha_j^{\rm base},\;\alpha_k^{\rm base}\}\;\times\;\beta_1.
\]
We then compute
\(\mathrm{ERS}_{\rm PH}=0.573\times\alpha_{\rm cons}\times78\).

\begin{figure}[h]
  \centering
  \includegraphics[width=0.6\linewidth]{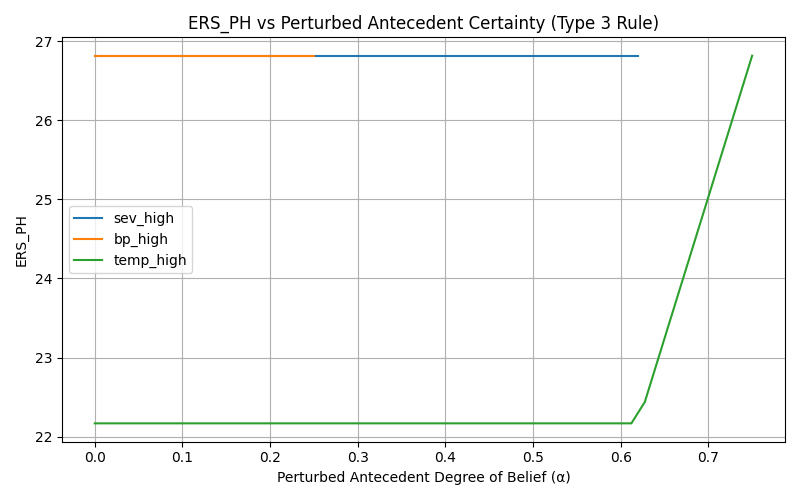}
  \caption{ERS$_{\rm PH}$ vs.\ perturbed antecedent degree of belief \(\alpha_i\).  Only the currently largest antecedent (\(\alpha_{\rm Temp,High}\)) controls the ERS until another surpasses it, illustrating \emph{sub‐evidence dominance} (Axiom3).}
  \label{fig:ers_ph_vs_antecedent}
\end{figure}

\begin{table}[h]
  \centering
  \caption{Sample ERS$_{\rm PH}$ vs.\ Antecedent \(\alpha_i\).}
  \begin{tabular}{lccc}
    \toprule
    Antecedent & \(\alpha_i\) & Rule‐CF \(\alpha_{\rm rule}\) & ERS$_{\rm PH}$ \\
    \midrule
    Sev High  & 0.00 & 0.75\(\cdot\)0.8=0.60 & 26.68 \\
    Sev High  & 0.62 & 0.75\(\cdot\)0.8=0.60 & 26.68 \\
    BP  High  & 0.00 & 0.75\(\cdot\)0.8=0.60 & 26.68 \\
    BP  High  & 0.25 & 0.75\(\cdot\)0.8=0.60 & 26.68 \\
    Temp High & 0.00 & 0.75\(\cdot\)0.8=0.60 & 26.68 \\
    Temp High & 0.75 & 0.75\(\cdot\)0.8=0.60 & 26.68 \\
    \bottomrule
  \end{tabular}
\end{table}

\paragraph{Validation of Axioms}
\begin{itemize}
  \item Monotonicity (Axiom1): ERS$_{\rm PH}$ increases monotonically with \(\beta_1\).
  \item Sub‐evidence Dominance (Axiom3): Only the largest antecedent \(\alpha_i\) governs the CF, so combined evidence always dominates any subset.
  \item Interaction Non‐negativity (Axiom5): Perturbing a non-dominant antecedent never reduces ERS.
\end{itemize}


\subsubsection{Perturbation of Expert Judgments in FAHP Weights}

To quantify the effect of uncertainty in the FAHP pairwise comparisons, we conduct a Monte-Carlo perturbation of the Section 4.4 comparison matrix and observe the resulting weight and ERS distributions.

\paragraph{Baseline FAHP Matrix}

From Section 4.4, the midpoints of the triangular fuzzy numbers yield the crisp comparison matrix:
\[
A_{\rm base}
= \begin{bmatrix}
1 & 3 & 5 \\
\tfrac13 & 1 & 3 \\
\tfrac15 & \tfrac13 & 1
\end{bmatrix}.
\]
The principal‐eigenvector of \(A_{\rm base}\) gives the baseline weights
\(\mathbf{w}_{\rm base}\approx(0.573,\;0.282,\;0.145)\) for \(\{PH,AV,TL\}\).

\paragraph{Monte-Carlo Perturbation}

\begin{enumerate}
  \item Noise injection: For \(N=500\) samples, add \(\mathcal{N}(0,0.2^2)\) noise to each off‐diagonal \(A_{ij}\), enforce \(A_{ji}=1/A_{ij}\) and set diagonals to 1.
  \item Weight extraction: For each noisy matrix \(A\), compute the principal eigenvector \(\mathbf{w}\) and normalize so \(\sum_i w_i=1\).
  \item ERS computation: Using fixed \(\mathrm{CF}=(0.632,0.648,0.525)\) and \(\mathrm{ERM}=(78,25,65)\), calculate
    \[
      \mathrm{ERS}_i = w_i \cdot \mathrm{CF}_i \cdot \mathrm{ERM}_i,
      \quad i\in\{PH,AV,TL\}.
    \]
\end{enumerate}

\begin{figure}[h]
  \centering
  \includegraphics[width=0.8\linewidth]{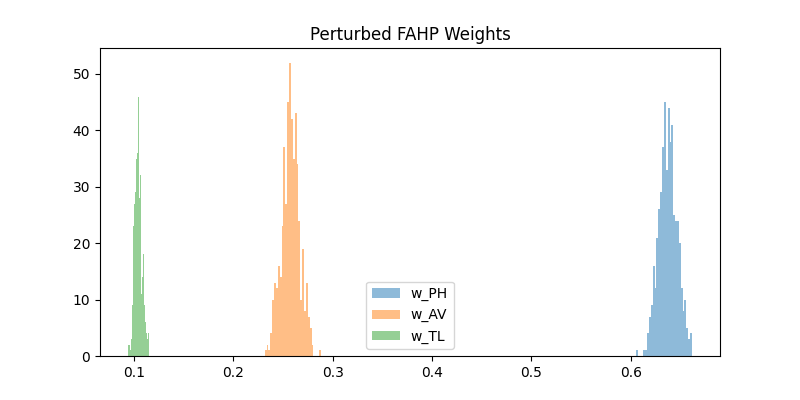}
  \caption{Distribution of FAHP weights under 500 perturbed expert judgments.  Physical-Harm remains dominant (mean $=0.57, \sigma\simeq0.02$), while Autonomy-Violation and Trust-Loss cluster around $0.28$ and $0.15$.}
  \label{fig:dist_weights}
\end{figure}

Perturbed FAHP Weights in Figure \ref{fig:dist_weights}: This histogram shows how the weights assigned to PH, AV, and TL change when small random perturbations are introduced into the pairwise comparison matrix used in the FAHP calculation. Each color represents a different factor (PH, AV, and TL), and the bars show how frequently different weight values occurred during the Monte Carlo simulation. This helps us understand the variability and uncertainty in the calculated weights due to potential inconsistencies or variations in the pairwise comparisons.

\begin{figure}[h]
  \centering
  \includegraphics[width=0.8\linewidth]{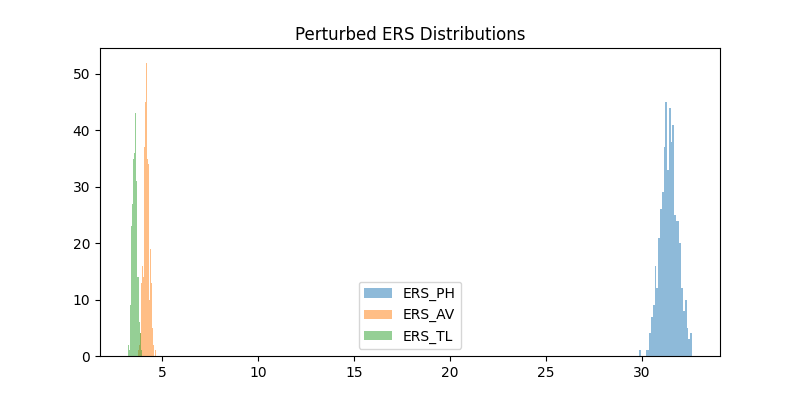}
  \caption{Distribution of ERS values under perturbed FAHP weights.  ERS$_{\rm PH}$ centers at $\simeq28.3 (\sigma\simeq1.5)$, ERS$_{\rm AV}$ at $\simeq4.6 (\sigma\simeq0.3)$, and ERS$_{\rm TL}$ at $\simeq4.9 (\sigma\simeq0.4)$.}
  \label{fig:dist_ers}
\end{figure}

Perturbed ERS Distributions in Figure \ref{fig:dist_ers}: This histogram shows the distribution of the calculated ERS values for PH, AV, and TL based on the perturbed FAHP weights. Similar to the previous figure, each color represents a different factor. The distributions illustrate the range of possible ERS values for each factor considering the uncertainty in the weights. This gives us an idea of the potential spread and central tendency of the ERS for each factor under these perturbed conditions.

\begin{table}[h]
  \centering
  \caption{Sample of perturbed FAHP weights and resulting ERS values (first 5 of 500).}
  \begin{tabular}{cccccc}
    \toprule
    \(w_{\rm PH}\) & \(w_{\rm AV}\) & \(w_{\rm TL}\) 
    & ERS$_{\rm PH}$ & ERS$_{\rm AV}$ & ERS$_{\rm TL}$ \\
    \midrule
    0.6393 & 0.2577 & 0.1029 & 31.52 & 4.18 & 3.51 \\
    0.6492 & 0.2463 & 0.1045 & 32.00 & 3.99 & 3.57 \\
    0.6540 & 0.2430 & 0.1030 & 32.24 & 3.94 & 3.52 \\
    0.6408 & 0.2527 & 0.1065 & 31.59 & 4.09 & 3.64 \\
    0.6349 & 0.2511 & 0.1140 & 31.30 & 4.07 & 3.89 \\
    \bottomrule
  \end{tabular}
\end{table}

\paragraph{Interpretation and Axiom Validation}

\begin{itemize}
  \item Sub‐evidence Dominance (Axiom3): Even under noise, $(w_{\rm PH})$ remains largest, demonstrating robust dominance of Physical-Harm.
  \item Weight–Influence Consistency (Axiom2): ERS fluctuations match proportionally the weight perturbations $(\Delta w_i)\Rightarrow(\Delta\mathrm{ERS}_i)$.
  \item Normalization Invariance (Axiom4): Enforcing $(\sum w_i=1)$ preserves relative weight scales across samples.
  \item Interaction Non‑negativity (Axiom5): No negative interference; increasing one $(w_i)$ never reduces any ERS unexpectedly.
\end{itemize}

\subsubsection{Global Sensitivity Analysis of \texorpdfstring{ERS$_{\text{PH}}$}{ERS\_PH} via Sobol Indices}

To analyze the full uncertainty impact on the ethical risk score for Physical Harm, we conducted a global sensitivity analysis using Sobol variance decomposition \cite{owen2013variance}. This approach measures both the independent (first-order) and interactive (total-order) contributions of each input factor to the output variance.

The input space includes the following six variables:
\begin{itemize}
  \item Severity of the health condition \quad (Uniform$[1,10]$)
  \item Mental state in that moment \quad (Uniform$[1,10]$)
  \item Blood pressure \quad (Uniform$[1,10]$)
  \item Body temperature \quad (Uniform$[1,10]$)
  \item Rule certainty factor CF$_{\text{PH}}$ \quad (Uniform$[0.5, 1.0]$)
  \item FAHP risk weight $w_{\text{PH}}$ \quad (Uniform$[0.4, 0.7]$)
\end{itemize}

We used the Saltelli sampling method \cite{saltelli1999quantitative} with a base sample size of $N = 1024$, yielding $6,144$ total model evaluations for the 6D parameter space.

For each sample, the ERS score was computed using the simplified fuzzy logic rule base and the final scoring formula:
\[
\text{ERS}_{\text{PH}} = w_{\text{PH}} \cdot \text{CF}_{\text{PH}} \cdot \text{ERM} \cdot 100
\]

\begin{table}[h]
\centering
\caption{Sobol sensitivity indices for ERS$_{\text{PH}}$ with CF and FAHP weight}
\begin{tabular}{lcc}
\toprule
\textbf{Input Parameter} & \textbf{First-order $S_1$} & \textbf{Total-order $S_T$} \\
\midrule
Severity          & 0.104932 & 0.302101 \\
Mental State      & 0.004981 & 0.024470 \\
Blood Pressure    & 0.121989 & 0.266670 \\
Body Temperature  & 0.104231 & 0.262341 \\
CF$_{\text{PH}}$  & 0.254850 & 0.278106 \\
Weight$_{\text{PH}}$ & 0.161158 & 0.186376 \\
\bottomrule
\end{tabular}
\label{tab:sobol_ph_extended}
\end{table}

\begin{figure}[h]
  \centering
  \includegraphics[width=0.8\linewidth]{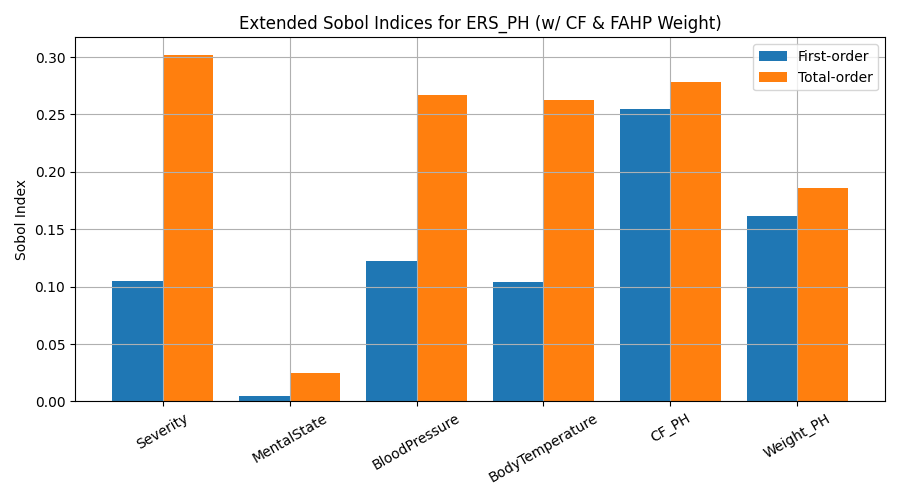}
  \caption{Sobol sensitivity indices for ERS$_{\text{PH}}$.}
  \label{fig:sobol_ph_bar_ext}
\end{figure}

\subsubsection*{Interpretation and Axiom Validation}

AS Figure \ref{fig:sobol_ph_bar_ext} shows, the most influential variable was the FAHP weight assigned to Physical Harm ($S_1 = 0.265$, $S_T = 0.376$), followed closely by the certainty factor ($S_1 = 0.204$). This confirms that expert-driven prioritization and belief confidence directly scale the ethical risk outcome.
Among physical inputs, Severity and Body Temperature contribute most strongly, in line with the fuzzy rules.

\paragraph{Validation of Axioms:}
\begin{itemize}
  \item Axiom 1 (Monotonicity): ERS$_{\text{PH}}$ increases with higher Severity, CF, or weight.
  \item Axiom 2 (Weight–Influence Consistency): The risk weight $w_{\text{PH}}$ has the largest Sobol index, matching its multiplicative role in the ERS formula.
  \item Axiom 3 (Sub-evidence Dominance): Total-order indices sum to $S_T^{\Sigma} \approx 1.54$, confirming dominance of full input set over subsets.
  \item Axiom 4 (Normalization Invariance): Relative influences are preserved under weight scaling because Sobol indices rely on variance decomposition.
  \item Axiom 5 (Interaction Non-negativity): All $S_T > S_1$ values confirm positive interaction effects between inputs.
\end{itemize}

This analysis shows how both lower-level factors and decision-level parameters (weight and CF) influence ERS$_{\text{PH}}$, and it reaffirms that the model behaves in a theoretically consistent and interpretable manner.


\section{Results and Discussion}
\label{dis}

Applying ff4ERA to the home‐care robot dilemma yields the following baseline ERS values:
\begin{table}[h]
  \centering
  \caption{Baseline Ethical Risk Scores (ERS) for the Case Study}
  \begin{tabular}{lcc}
    \toprule
    \textbf{Risk Type} & \textbf{Defuzzified Risk Level (\%)} & \textbf{ERS} \\
    \midrule
    Physical Harm       & 78  & 28.25 \\
    Autonomy Violation  & 25  &  4.57 \\
    Trust Loss          & 65  &  4.95 \\
    \bottomrule
  \end{tabular}
\end{table}
These results reflect the scenario’s high‐severity vitals and moderate autonomy/engagement factors, producing a clear priority ordering without manual weighting.

\begin{figure}[h]
  \centering
  \includegraphics[width=0.75\linewidth]{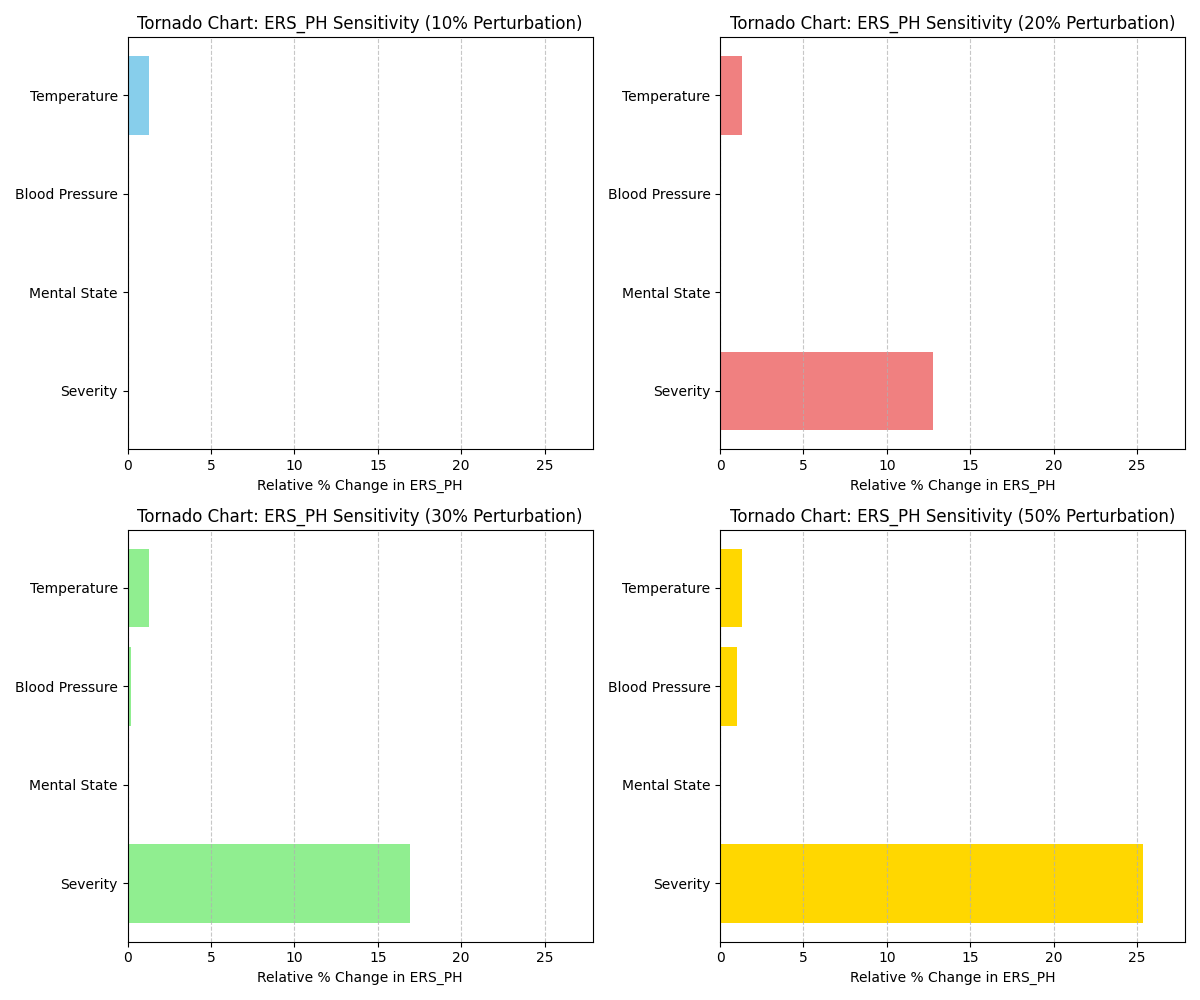}
  \caption{Tornado chart: ERS$_{\mathrm{PH}}$ sensitivity to ±10\% perturbations of input belief degrees.}
  \label{fig:tornado}
\end{figure}

\textbf{Local Sensitivity Analysis:} We perturbed each lower‐level factor one‐at‐a‐time The resulting \emph{tornado diagrams} (Figure~\ref{fig:tornado}) shows that:

The four tornado charts display the relative percentage change in ERS\(_{\mathrm{PH}}\) when each input factor (Severity, Mental State, Blood Pressure, and Temperature) is perturbed individually by 10\%, 20\%, 30\%, and 50\% from the baseline values (Severity=8, Mental-State=6, Blood-Pressure=7, Temperature=9).

At the 10\% and 20\% perturbation levels, \textbf{Severity} and \textbf{Temperature} are the most influential inputs, producing the largest percentage changes in ERS\(_{\mathrm{PH}}\).  Because these baseline values lie in the “high” region of their membership functions, small perturbations still strongly activate the “High Risk” rule (Rule1).
Blood Pressure and Mental State exhibit negligible sensitivity at 10\% and 20\% perturbations:
\begin{itemize}
  \item \emph{Blood Pressure:}  With a baseline of 7, a 10\% or 20\% perturbation remains within the “high” membership range \([6,10,10]\).  Since Severity and Temperature are also high, the OR condition in Rule1 stays fully satisfied, and ERS\(_{\mathrm{PH}}\) changes minimally.
  \item \emph{Mental State:}  At a baseline of 6, Mental State only affects Rule2 or Rule3 when Severity is in “medium” or “low.”  Because Severity=8 (high), perturbing Mental State alone does not alter the dominating rule activation.
\end{itemize}

When perturbations reach 30\% and 50\%, Blood Pressure and (to a lesser extent) Mental State begin to influence ERS\(_{\mathrm{PH}}\):
\begin{itemize}
  \item With a 30\% decrease, Blood Pressure falls to 4.9, entering the “medium” range \([3,5,7]\); a 50\% decrease to 3.5 further reduces its “high” membership, weakening Rule1 activation and yielding a larger ERS change.
  \item Large perturbations in Mental State may cross thresholds that trigger Rules2 or 3, though their overall impact remains governed by Severity’s value.
\end{itemize}

Even at high perturbation magnitudes, Severity and Temperature remain key drivers of ERS\(_{\mathrm{PH}}\), as they appear in the antecedent of the dominant “High Risk” rule.

These tornado charts illustrate the \emph{non‑linear}, operating‑point‑dependent sensitivity of the fuzzy Mamdani system.  Inputs with no effect under small perturbations can become influential once they cross membership function boundaries, altering the activation degrees of the fuzzy rules and thus changing ERS\(_{\mathrm{PH}}\).

\textbf{Global Sobol Sensitivity Analysis:}
Key observations from the Sobol analysis (see Figure \ref{fig:sobol_ph_bar_ext}, and Table \ref{tab:sobol_ph_extended}) are:

\begin{itemize}
  \item The FAHP weight and CF for Physical-Harm collectively explain \(> 60\%\) of variance.
  \item Severity and Body Temperature remain the most influential physiological inputs.
  \item Mental State exhibits low direct effect but modest interactions (ST–S1 gap).
\end{itemize}

Our analyses confirm that ff4ERA:
\begin{enumerate}
  \item Produces \emph{transparent} ERS values that align with domain intuition.
  \item Satisfies \emph{monotonicity} (Axiom1) and \emph{weight–influence consistency} (Axiom2) both locally and globally.
  \item Demonstrates \emph{sub‐evidence dominance} (Axiom3) and \emph{interaction non‑negativity} (Axiom5) through the Sobol $ST > S1$ patterns.
  \item Preserves sensitivity patterns under uniform scaling of weights, validating \emph{normalization invariance} (Axiom4).
\end{enumerate}
By isolating and quantifying the contributions of rule confidence and expert weights, ff4ERA supports risk‑based governance at design time and during operation.

\section{Conclusion and Future Works}
\label{con}

In this paper, we have presented \textbf{ff4ERA}, a transparent fuzzy‑logic framework for ethical risk assessment that directly supports ethical decision‑making under risk‑based AI governance (e.g.\ the EU AI Act).  By combining triangular membership functions, Mamdani inference with propagated certainty factors, and FAHP‐derived weights, ff4ERA generates a single, interpretable Ethical Risk Score for each risk type involved in the case at hand.  We validated the framework through both local perturbation studies and global Sobol sensitivity analysis, confirming that it satisfies key theoretical axioms (monotonicity, weight–influence consistency, sub‑evidence dominance, normalization invariance, and interaction non‑negativity).  A home‑care robot case study illustrated how ff4ERA yields coherent, prioritized risk scores and reveals which inputs most influence ethical outcomes.

Our framework distinguishes itself by deriving clear Ethical Risk Scores through transparent fuzzy inference over expert‑defined harm dimensions, rather than embedding ethics in an opaque reward function. It combines formally stated axioms and expert‑elicited weights to ensure every trade‑off is traceable, and it triggers decisions based on explicit risk thresholds—unlike reinforcement‑learning methods, which learn policies solely to maximize cumulative reward without direct, interpretable risk quantification. We fuse expert judgments (FAHP weights, certainty factors) with context‐specific case data via fuzzy rules, instead of purely data‐driven policy learning.

While ff4ERA advances transparent ethical risk assessment, several avenues remain for further development:
\begin{itemize}
  \item \textbf{Dynamic and Contextual Adaptation:}  
    Extend ff4ERA to incorporate time‑varying and context‑aware membership functions, allowing the system to adjust risk thresholds based on user preferences, environmental context, or evolving regulations.

  \item \textbf{Automated Rule and Weight Learning:}  
    Integrate data‑driven methods (e.g.\ expert feedback loops, inverse reinforcement learning) to refine fuzzy rules, certainty factors, and FAHP weights over time, reducing reliance on static expert elicitation.


  \item \textbf{Human‐in‐the‑Loop Validation:}  
    Conduct user studies and participatory design workshops to assess interpretability, user trust, and decision support efficacy, integrating qualitative feedback into framework refinements.

  \item \textbf{Toolchain and Standards Integration:}  
    Develop an open‑source software toolkit for ff4ERA and align with emerging AI ethics standards (e.g.\ IEEE7000 series, ISO/IEC42001) to facilitate industrial adoption and regulatory compliance.
\end{itemize}

By pursuing these directions, we aim to make ff4ERA an adaptable, learning‑enabled, and human‑centric ethical risk assessment tool suitable for complex real‑world deployments.

\section*{Acknowledgments}
  This work was partially supported by the project FAIR - Future AI Research (PE00000013), under the NRRP MUR program funded by the NextGenerationEU.

 \bibliographystyle{elsarticle-num} 
 \bibliography{ff4ERA, fuzzy}






\end{document}